\documentclass{article}

\usepackage{arxiv}

\usepackage[utf8]{inputenc} 
\usepackage[T1]{fontenc}    
\usepackage{lmodern}
\usepackage{hyperref}

\usepackage{url}            
\usepackage{booktabs}       
\usepackage{amsfonts}       
\usepackage{nicefrac}       
\usepackage{microtype}      
\usepackage{lipsum}

\usepackage{rotating}
\usepackage{graphicx}
\usepackage{subcaption}
\usepackage{tabularx,ragged2e}
\usepackage{array}
\usepackage{amsmath}
\usepackage{amsfonts}
\usepackage{amssymb}
\usepackage{graphicx}
\usepackage{multirow}
\usepackage{array}
\usepackage{csquotes}
\usepackage{xcolor}
\definecolor{MyGreen}{RGB}{0, 153, 0}
\definecolor{MyOrange}{RGB}{204, 102, 0}
\definecolor{MyPurple}{RGB}{204, 0, 102}
\usepackage[linesnumbered,lined,boxed,commentsnumbered]{algorithm2e}

\usepackage{float}
\usepackage{algorithmic}
\usepackage{longtable}
\usepackage{rotating}
\usepackage{multicol}
\usepackage{booktabs}
\usepackage{bm}

\usepackage{booktabs}

\title{DENS-ECG: A Deep Learning Approach for ECG Signal Delineation}

\author{
  Abdolrahman Peimankar\thanks{A. Peimankar can also be contacted at: \texttt{a.peimankar@gmail.com}} \\
  Department of Health Technology\\
  Technical University of Denmark\\
  2800 Kgs. Lyngby, Denmark \\
  \texttt{apeima@dtu.dk} \\
   \And
 Sadasivan Puthusserypady \\
  Department of Health Technology\\
  Technical University of Denmark\\
  2800 Kgs. Lyngby, Denmark \\
  \texttt{sapu@dtu.dk} \\
}

\begin{document}
\maketitle

\begin{abstract}
\textit{Objectives:} With the technological advancements in the field of tele-health monitoring, it is now possible to gather huge amounts of electro-physiological signals such as electrocardiogram (ECG). It is therefore necessary to develop models/algorithms that are capable of analysing these massive amounts of data in real-time. This paper proposes a deep learning model for real-time segmentation of heartbeats. \textit{Methods:} The proposed algorithm, named as the DENS-ECG algorithm, combines convolutional neural network (CNN) and long short-term memory (LSTM) model to detect onset, peak, and offset of different heartbeat waveforms such as the P-wave, QRS complex, T-wave, and No wave (NW). Using ECG as the inputs, the model learns to extract high level features through the training process, which, unlike other classical machine learning based methods, eliminates the feature engineering step. \textit{Results:} The proposed DENS-ECG model was trained and validated on a dataset with 105 ECGs of length 15 minutes each and  achieved an average sensitivity and precision of 97.95\% and 95.68\%, respectively, using a 5-fold cross validation. Additionally, the model was evaluated on an unseen dataset to examine its robustness in QRS detection, which resulted in a sensitivity of 99.61\% and precision of 99.52\%. 
\textit{Conclusion:} The empirical results show the flexibility and accuracy of the combined CNN-LSTM model for ECG signal delineation. \textit{Significance:} This paper proposes an efficient and easy to use approach using deep learning for heartbeat segmentation, which could potentially be used in real-time tele-health monitoring systems. 
\end{abstract}


\section{Introduction}
\label{sec1}
Analysis of electrocardiogram (ECG) signals is one of the most important steps in the diagnosis of cardiac disorders. In order to achieve high diagnostic accuracies, almost all the ECG  analysis tools/software require the knowledge about the location and morphology of different segment  waveforms (P-QRS-T) in  ECG records. For example, atrial fibrillation (AFIB) is one of the most common cardiac arrhythmias in elderly population \cite{r3,r4} and P-wave absence is one of the important and clinically useful features for the detection of AFIB \cite{r1,r2}. This makes P-wave delineation of great importance in clinical practice. In addition, most of the developed state-of-the-art algorithms for analysing ECG records are also dependant on the detection of QRS complexes (R-peaks) \cite{r5,r6,ra7,r7,ALONSOATIENZA2012,hagiwara2018computer,KHALAF2015,CEYLAN2009,luz2013ecg,HOMAEINEZHAD2012}. These algorithms use extracted RR intervals to define various features, which can be used for classification of different arrhytmias. It should be noted that these methods usually utilise the annotated databases such as the Physionet datasets to validate their performance \cite{r8}. For example, \cite{a1,a2,a3,a4,a5,a6,MARTIS2012,KHORRAMI2010} proposed algorithms using RR intervals based features from the annotated databases to detect cardiac arrhythmias. However, these methods become extremely cumbersome to test on the newly collected raw ECG signals in real-world applications.              

The latest advancements in tele-health monitoring systems provide the opportunity to collect huge amounts of ECG data. One of the most common ways for physicians/cardiologists to analyse ECG waveforms is through visual examination of the recordings. However, it is not always easy and in most cases difficult as well as extremely time consuming to analyse such huge amounts of data. Subjectivity is another big concern in such approaches. Therefore, it is of great interest to develop a reliable software for ECG signal delineation in order to specify the accurate location of different segment waves, such as the P-wave, QRS complex, and T-wave, for subsequent use in diagnostics. 

Various state-of-the-art algorithms for ECG delineation have been introduced in the literature. Most of these algorithms are based on classical machine learning and digital signal processing techniques. \cite{r43} proposed a wavelet based ECG delineation algorithm. This algorithm was validated on different datasets achieving very high performance. \cite{r9} used generalised orthogonal forward regression with Gaussian mesa function models for automatic ECG wave extraction. \cite{r11} proposed a Bayesian model for P- and T- waves detection, which showed higher accuracy compared to previously published algorithms but at a higher computational cost. 


Classical machine learning methods are required to define specific features from the data and use them as inputs in order to train the models. Feature engineering is an essential step to transform the raw data into a suitable internal representation from which the model is able to distinguish between the different classes \cite{r23}. On the other hand, deep learning models overcome this limitation by automatically extracting the relevant features directly from the data and outperforms many state-of-the-art models within different fields and applications of machine learning. This is because, deep learning models are capable of extracting highly abstract features from signals without the need for prior domain knowledge and expertise \cite{r23,r24}. 

The applications of deep learning in tele-health and biomedical engineering is growing exponentially in recent years. These methods have been used to solve many challenges in biomedical signal processing applications. For example, these methods were applied successfully to ECG arrhythmias classification \cite{r44,r45,r46,r47} and electroencephalogram (EEG) signal classification \cite{r48,r49,r50}. It has been shown in the literature that a combination of convolutional neural networks (CNN) and long short-term memory (LSTM) can enhance the classification/prediction performance as both the CNN and LSTM learn different (complex) functions from the input signals in the training phase\cite{r25,r26,r27,r28,r29}.  In this paper, a deep combined CNN-LSTM model, named as the DENS-ECG algorithm, is proposed to automatically extract features from ECG records. These features are subsequently used to distinguish between four distinct component waveforms (i.e. P-wave, QRS complex, and T-wave) in each heartbeats.

The remainder of this paper consists of 4 sections. In Section \ref{sec2}, the background of the deep learning model and the proposed DENS-ECG algorithm for ECG signal delineation are briefly described. The experimental results are presented in Section \ref{sec3}. Section \ref{sec4} presents the discussion and comparison with other state-of-the-art methods, followed by the conclusions in Section \ref{sec5}. 

\section{Materials and Methods}\label{sec2}
\subsection{Dataset}
In this study, PhysioNet QT database (QTDB) was used to train and validate the performance of the proposed algorithm \cite{r21}. There are 105 records in the QTDB and each has a length of 15 minutes with a sampling frequency of 250 Hz. In addition, the MIT-BIH Arrhythmia Database (MITDB) was used to test the model. It contains 48 half-hour ECG recordings, which were sampled at 360 Hz \cite{r42}. PyhsioNet's WFDB python package \cite{r8} was used to read the signals and their corresponding annotations.

\subsection{Pre-processing}
First, each record was filtered and segmented accordingly. The signals were zero-phase filtered using a 3rd order  Butterworth band-pass ($0.5-40$Hz) filter to remove baseline wanders and high frequency noises \cite{r22}. The records were then segmented into smaller chunks of 1000 samples. It should be emphasized that 84 out of 105 records of QTDB were used for training the model and the remaining 26 records were used for evaluating the model.

\subsection{Deep Learning Model Structure}
       
As discussed, the model used in this study for ECG signal delineation is a combination of two well-known deep network structures. The first part of the model consists of three 1D convolutional layers, which extracts the high abstract features from ECG segments. The second part of the model includes two deep LSTM layers to process the features extracted by the convolutional layers. Lastly, the output of the second LSTM layer is passed through a dense layer with four neurons, which provide the posterior probabilities corresponding to each of the four classes. It is worth noting that the dense layer is a time distributed layer to keep the continuity of the ECG records. 

\subsubsection{CNN layer}
\begin{figure}[htbp]
  \centering
  \includegraphics[width=0.5\linewidth]{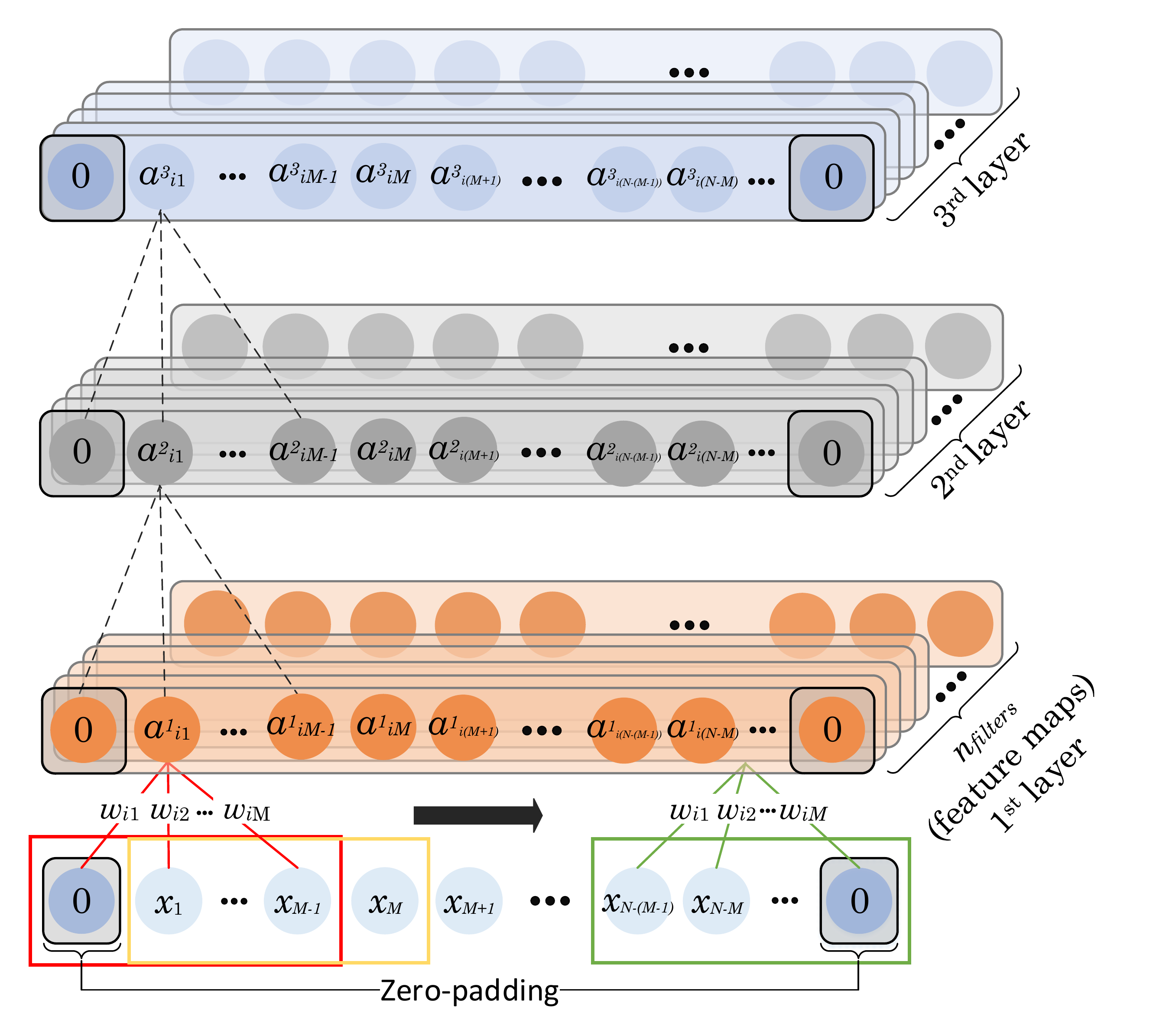}
\caption{Schematic diagram of the convolution process. A kernel of size $k_{size} = M$ moves across the input signal. The corresponding weights ($w_{i1}, \ldots, w_{i2}, w_{iM}$) are fixed for all the convolution operation. The signals in each layer are appropriately zero-padded to keep the dimensions same as in the input layer. For example, if kernel size ($k_{size}$) $M$ is odd, $(M-1)/2$ zeros are padded to each end of the signal, otherwise the zero-padding size is $M/2$.}
\label{conv}
\end{figure}

Unlike traditional neural networks, in which each neuron is connected to every neuron in the adjacent layer, CNNs are able to exploit any existing spatial and temporal patterns in the data \cite{r23}. For this purpose, CNNs take advantage of four key attributes, which are: 1) establishing local connections; 2) shared weights; 3) very large number of layers/filters; and 4) reducing the complexity of the network \cite{r23}. For example, in 1D CNNs, different filters are defined by sliding a fixed window over the signal. The length of the window used in CNNs for the convolution process is known as the kernel size, denoted as $k_{size}$. The outputs of these convolutions (between the filters and specific regions of the input signal) are the neurons in the resulted feature maps as illustrated in Figure \ref{conv}. The weights of these connections and an overall bias are learned during the training process. It should be noted that there is only one set of weights corresponding to each feature map as shown in Figure \ref{conv}. This convolution process can be expressed as follows \cite{r30}:
\begin{equation}\label{eq1}
a_{ij}^m = \varphi\bigg(b_i + \sum_{k=1}^M w_{ik}x_{j+k-1}\bigg) = \varphi(b_i + \mathbf{w}_i^T\mathbf{x}_j),
\end{equation}
\noindent where $a_{ij}^m$ is the activation or the output of the $j$th neuron of the $i$th filter for the $m$th convolutional layer, $M$ is the kernel size, $\varphi$ is the neural activation function, $b_i$ is the shared bias of the $i$th filter, $\mathbf{w}_i = [w_{i1}\quad w_{i2}\quad \ldots\quad w_{iM}]^T$ are the shared weights of the $i$th filter, and $\mathbf{x}_j =[x_j\quad x_{j+1}\quad\ldots\quad x_{j+M-1}]^T$ are the corresponding $M$ inputs. The outputs of the neurons, $a$'s, are the filtered version of the input time series, which learns the same features at different locations as the filter (kernel) slides over the input signal. Applying various filters to the input time series leads to different feature maps in the output of the activation functions. As shown in Figure \ref{conv}, the number of feature maps are equal to the number of applied filters ($n_{filters}$), which are each defined by a set of $M$ shared weights and a single bias (the biases $b_i$'s are not shown in the figure). 

\subsubsection{LSTM layer}
\begin{figure}[htbp]
  \centering
  \includegraphics[width=0.5\linewidth]{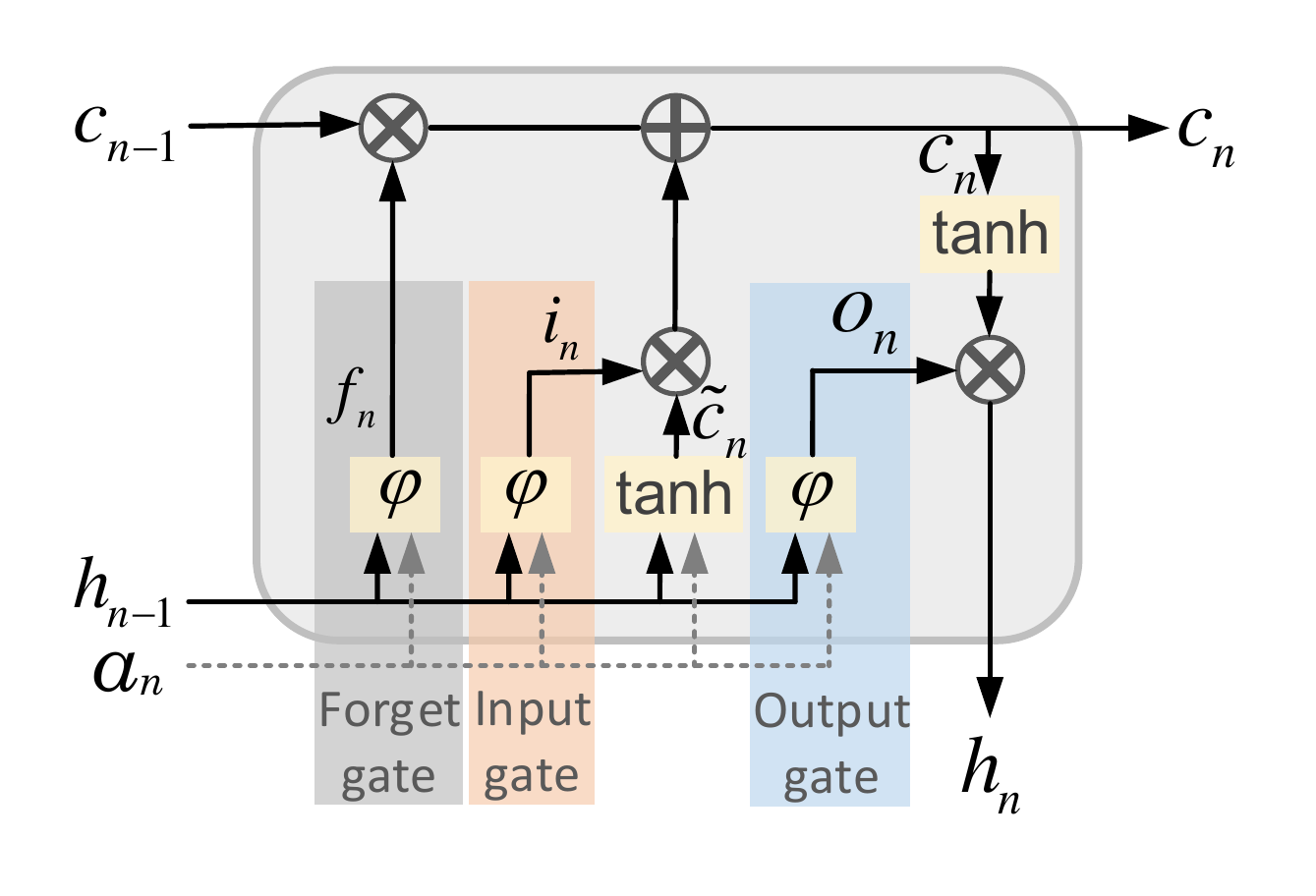}
\caption{Schematic diagram of LSTM memory block.}
\label{lstm}
\end{figure}

Recurrent Neural Networks (RNNs) are designed to work with sequential time-series which are capable of learning dependencies in sequential information. However, it has been shown that learning long-term dependencies are very challenging \cite{r31}. LSTM networks, a special type of RNNs, are capable of addressing the problem of unstable gradient and can handle long-term dependencies \cite{r32}. As depicted in Figure \ref{lstm}, there are three main parts in a LSTM block: (i) forget gate ($f_n$), (ii) input gate ($i_n$), and (iii) output gate ($o_n$). Forget and output gates are mainly responsible to remove or add information to the memory block in the following way:
\begin{align}\label{eq2}
f_{n} = \varphi(b_f + \mathbf{u}_{f}^T \mathbf{a}_n + \mathbf{w}_{f}^T \mathbf{h}_{n-1}),\\
i_n = \varphi(b_{i} +  \mathbf{u}_{i}^T  \mathbf{a}_n +  \mathbf{w}_{i}^T \mathbf{h}_{n-1}),
\end{align}
\noindent where $\mathbf{a}_{n}$ is the input sequence to the LSTM at time step $n$, which is actually the output of the last CNN layer here, and $\mathbf{h}_{n-1}$ is the output sequence at time step $n-1$. The $\mathbf{u}_{f}$, $\mathbf{w}_{f}$, $\mathbf{u}_{i}$, and $\mathbf{w}_{i}$ represent the weight vectors and $b_f$ and $b_i$ are bias terms. These  should be learned in the training phase of the LSTM. In addition, since $0\leq\varphi(\cdot)\leq1$, this controls the contribution of each unit in the memory block. Therefore, the memory $c_n$ is updated as:
\begin{equation}\label{eq4}
c_n = f_n c_{n-1} + i_n \tilde{c}_n,
\end{equation}
where
\begin{equation}\label{eq5}
\tilde{c}_n = \tanh(b_c + \mathbf{u}_{c}^T \mathbf{a}_n + \mathbf{w}_{c}^T \mathbf{h}_{n-1}).
\end{equation}
Finally, the output vector $h_{n}$ is computed as:
\begin{equation}\label{eq6}
h_n = o_n \tanh(c_n), 
\end{equation}
where
\begin{equation}\label{eq7}
o_n = \varphi (b_o + \mathbf{w}_o^T \mathbf{a}_n + \mathbf{u}_o^T \mathbf{h}_{n-1}).
\end{equation}
Here, $\mathbf{u}_o$ and $\mathbf{w}_o$ are the weight vectors of the output gate, and $b_o$ is the output bias. From \eqref{eq6} and \eqref{eq7}, in addition to input and previous output gate, the current memory plays an important role in the output gate. This provides LSTM with the ability to keep or forget the existing memory efficiently \cite{r33}. 

Bidirectional LSTM (BiLSTM) is a variant of the LSTM, which can process a sequence of data in both directions. Unlike LSTM, BiLSTM can also exploit the future context \cite{r34}. It consists of two hidden layers, which are fed forward to the output layer \cite{r34}. The outputs of BiLSTM are a function of forward and backward pass along with their corresponding weights and biases. 

\subsubsection{Model training}
In order to find the optimum network parameters (e.g. weights and biases) and consequently achieve the optimal performance, the network needs to be trained appropriately. It is a non-convex optimisation problem, which can be solved by using a cost function ($\mathcal{J}$) in an iterative process \cite{r30,r35,r36}. One of the most commonly used cost function for multi-class classification problems is the categorical cross-entropy loss (CCEL), which consists of a softmax and a cross-entropy loss (CEL). The CCEL is utilised to output a probability over the different classes. For this purpose, the class labels are one-hot encoded, which converts all the elements of the label vector into zero except for the true class. The CEL function can be formulated as follows:
\begin{equation}\label{eq8}
\mathcal{J}_{CEL} = -\displaystyle\sum_{i=1}^{N_c} y_i \log(z_i),
\end{equation}
where $z_i$ is the computed score from the network corresponding to the true class, ${N_c}$ is the number of classes, and $y_i$ is non-zero only for the true class. Thus, the CCEL can be calculated as:
\begin{equation}\label{eq10}
\mathcal{J}_{CCEL} = -\displaystyle\sum_{i=1}^{N_c}\log(\sigma(\mathbf{z})_i),
\end{equation}
where
\begin{equation}\label{eq9}
\sigma(\mathbf{z})_i = softmax = \dfrac{e^{z_i}}{\sum_{j=1}^{N_c}{e^{z_j}}}, \quad i = 1, \ldots, {N_c}.
\end{equation}

\subsection{Classification}
The output of the LSTM layer, which are the extracted features from ECG signals, is fed into a \textit{TimeDistributed} dense layer with four neurons with \textit{softmax} activation functions. The latter ensures a classification per time stamp such that the sum of the neuron outputs is equal to 1, i.e., they can be interpreted as posterior probabilities. The output of the dense layer for the \textit{i}$^{th}$ sample is classified to one of the four classes (P, QRS, T, or NW) as follows:
\begin{equation}\label{eq18}
\hat{y}_i = c_j \quad \mbox{if} \quad P(y_i=c_j|x_i) =
\operatorname*{argmax} P(c_1,\ldots,c_k|x_i),
\end{equation}

\noindent where $\hat{y}_i$ is the predicted class for the \textit{i}$^{th}$ sample $x_i$ and $P$ represents the posterior probability.

\subsection{Deep ECG Delineation Framework}\label{sec3}
The proposed deep learning model utilizes the advantages of ensemble learning technique to delineate ECG signals. The flowchart for the proposed DENS-ECG algorithm is illustrated in Figure \ref{flowchart}, which is described step by step as follows:

\begin{figure}[htbp]
  \centering
  \includegraphics[width=0.8\linewidth]{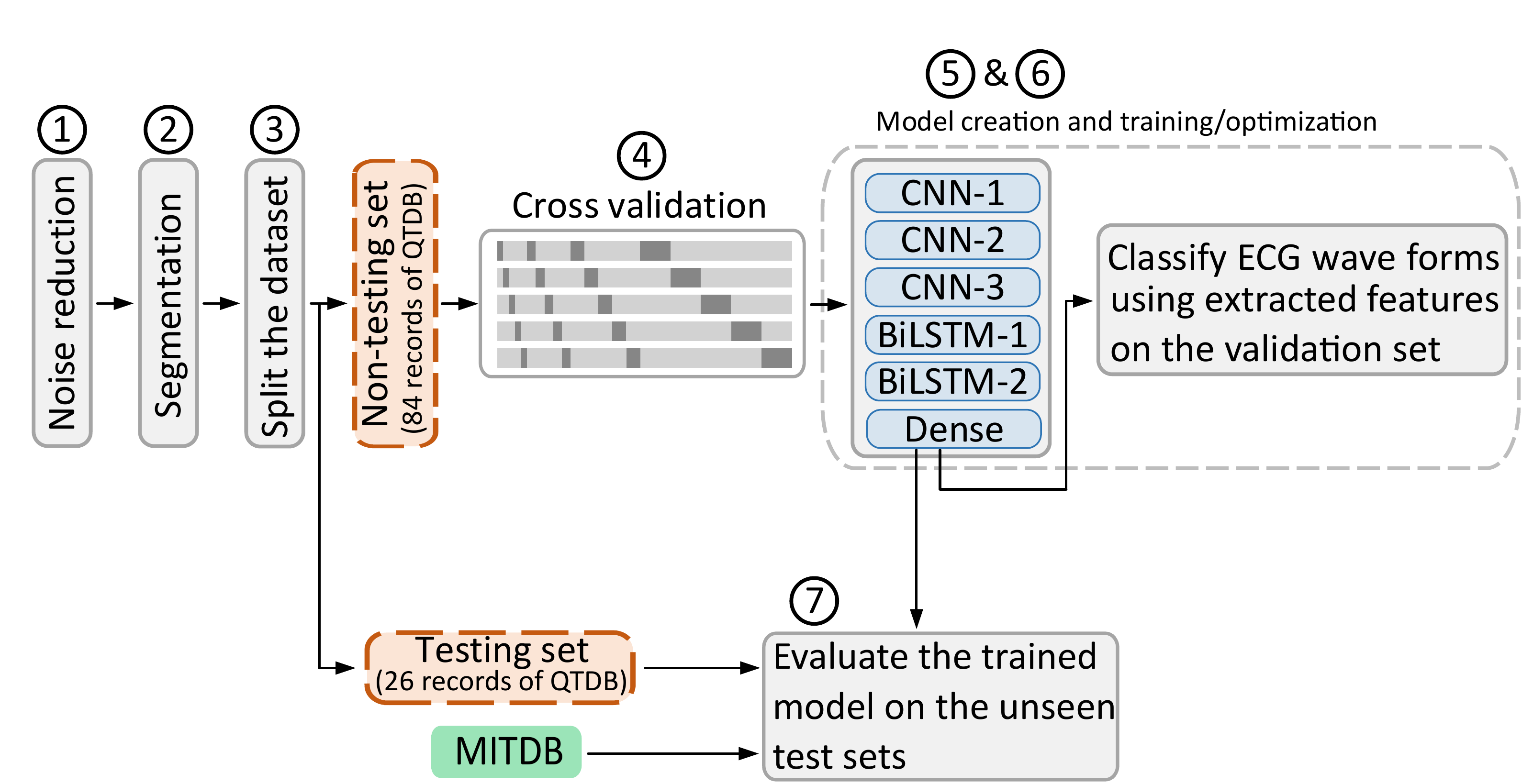}
\caption{Flowchart of the proposed DENS-ECG algorithm.}
\label{flowchart}
\end{figure} 

\begin{enumerate}
\item \textit{Noise reduction}: The ECG signals are filtered to remove noise and baseline wanders. 
\item \textit{Segmentation}: In this step, the ECG signals are segmented into chunks of 1000 samples. These segments are then fed into the model as inputs. It should be noted that the continuity of the time series (ECG signal) within each segment is preserved to make it possible for the network to learn the pattern of different waveforms from each input. 
\item \textit{Separate the testing set from a non-testing set}: The segmented ECG signals are divided into two sets. The non-testing set is used to train, validate and optimise the P-QRS-T waveforms delineation algorithm and the testing set is considered to evaluate the proposed model. The records in testing and non-testing sets are unique, i.e. no excerpt of the test records is included in the training process of the model. 

\item \textit{Cross validation}: The model is trained using 5-fold cross validation technique \cite{r38}. A stratified 5-fold cross validation (5-fold CV, Figure \ref{StratCV}) is used, where the distribution of samples in each fold is proportional to the size of the corresponding classes in the whole dataset. According to \cite{r39}, this method of  cross validation leads to a more reliable performance in terms of bias and variance compared to traditional cross validation techniques.

\begin{figure}[htbp]
  \centering
  \includegraphics[width=0.7\linewidth]{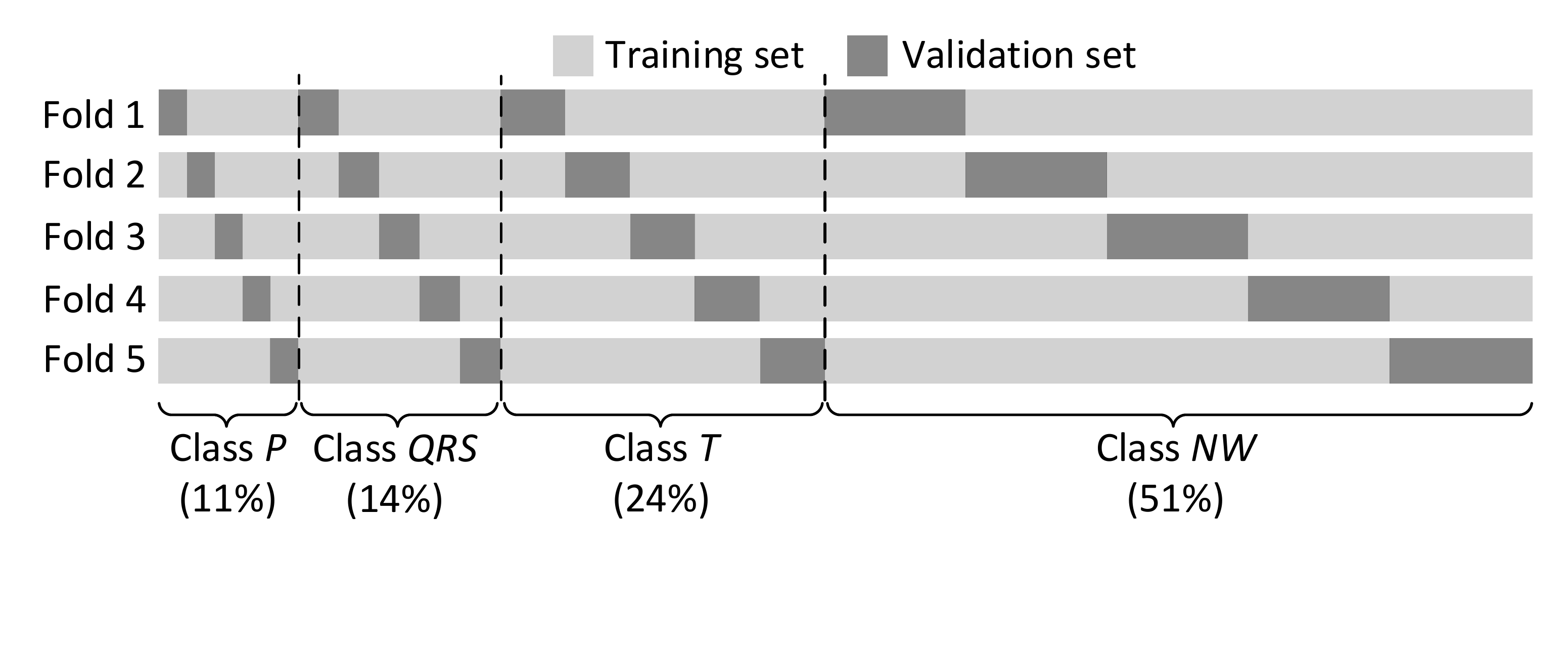}
\caption{Schematic diagram of a stratified 5-fold cross validation technique. The training and validation sets are split according to the size of the four classes.}
\label{StratCV}
\end{figure}

\item \textit{Creating the model}: In total, the model has eight layers, which includes the input layer, three 1D  convolutional layers followed by two BiLSTM and a dropout layer. Finally, a time distributed wrapper is used for the dense layer, which configures the BiLSTM layer for the sequence prediction. The input segments are fed directly into three successive convolutional layers, which extract the temporal patterns (features) from the ECG signals. A kernel size of $M=3$ is applied in the three convolutional layers and the corresponding number of filters ($n_{filters}$) are respectively 32, 64, and 128 for the three successive layers. In order to keep the same dimension in the input and  convolution layers, zero padding is employed. For example, the output of the first convolutional layer is now a sequence of 32 features with the same dimension of the input signal (time series). This process is repeated for the other two layers. The output of the last convolutional layer is 128 highly abstract feature maps, which are used as inputs for the first BiLSTM layer with $n_{units} = 250$  hidden units. The second BiLSTM layer has $n_{units} = 125$ hidden units. The dropout probability in the dropout layer is set to 0.2. The dropout layer helps to avoid over-fitting problem during the training of the network. The dense layer has 4 hidden units and a \textit{softmax} function is used as an activation function, which assigns a value between 0 and 1 to each sample of the input ECG signals. 
\item \textit{Model training and optimisation}: The model is trained using Adam optimisation algorithm \cite{r35}, which is different from the steepest gradient descent (SGD) optimisation algorithm. Adam is used for solving non-convex optimisation problems and is well-suited for large scale network. It has four hyper-parameters, which require to be fine-tuned; 1) the learning rate, $\alpha$, 2) the exponential decay rate for the first moment estimates, $\beta_1$, 3) the exponential decay rate for the second-moment estimates, $\beta_2$, and 4) the numerical stability parameter, $\epsilon$. A random search technique is used to find the optimum values of these parameters \cite{r37}. This approach is shown to be more efficient compared to the  grid search method for hyperparameters optimisation. The proposed DENS-ECG model contains 1,416,044 trainable parameters (weights and biases), which need to be optimised through the learning process. The model is implemented in Python 3.6.4 using \textit{keras} API \cite{a8}. Figures \ref{fig1:sfig1} and \ref{fig1:sfig2} show the loss and accuracy of the model during the training phase, respectively. Moreover, an early stopping technique is used to prevent over-fitting of the model. The model stops training if there is no decrease in the value of validation loss after three epochs. Additionally, the dropout process is only applied during the training phase of the model. Therefore, the training becomes more challenging for the network which in turn alleviate over-fitting problem. As illustrated in Figures \ref{fig1:sfig1} and \ref{fig1:sfig2},  the model achieves higher performance on the validation set than train data, which shows that the model is trained properly without over-fitting.

\begin{figure}[t]\centering
\begin{subfigure}{0.45\textwidth}
  \centering
  \includegraphics[width=0.9\linewidth]{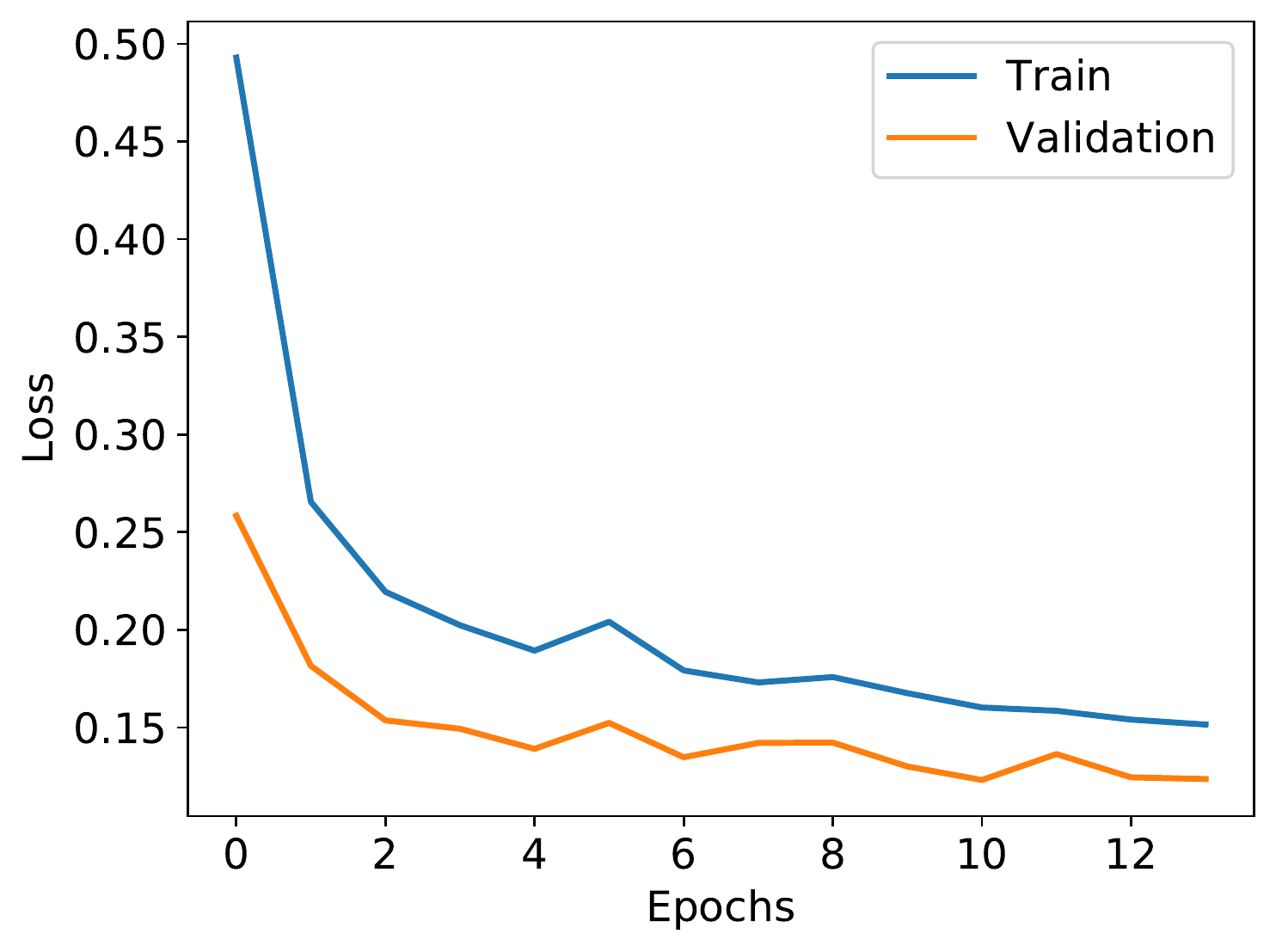}
  \caption{}
  \label{fig1:sfig1}
\end{subfigure}
\begin{subfigure}{0.45\textwidth}
  \centering
  \includegraphics[width=0.9\linewidth]{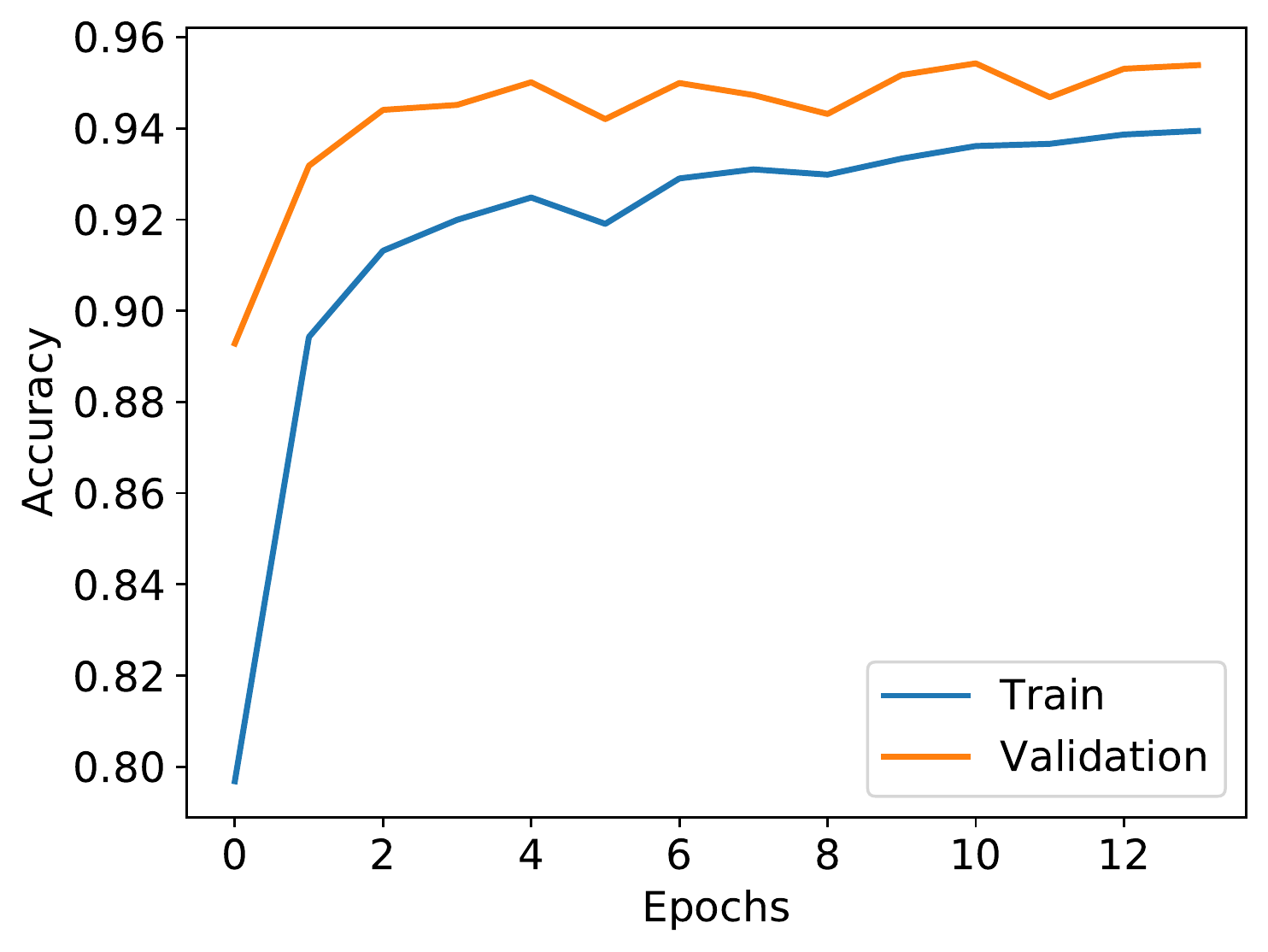}
  \caption{}
  \label{fig1:sfig2}
\end{subfigure}
\caption{Training and validation curves of the model for: (a) loss and (b) accuracy.}
\label{fig1}
\end{figure}

\item \textit{Evaluate the trained model}: The trained model is then evaluated on the 26 unseen test records from QTDB dataset to examine the performance of the classifier. In addition, the model is tested on the unseen MITDB dataset for QRS detection.

\end{enumerate}

\section{Results}\label{sec3}
In this study, we used the QTDB from PhysioNet  to develop the model \cite{r21}. In total, there are 105 ECG records of which 84 records (80\%) are used for training/validation and the remaining 26 records (20\%)  for testing the model. In addition, the robustness of the proposed model is examined using a different dataset, the MITDB. Since there is only QRS peak annotation available for MITDB, the performance of the proposed DENS-ECG model is reported on detecting the QRS complexes for this dataset.

\subsection{Classification Performance Metrics}\label{sec4}
The key factor in evaluating the performance of any classification system is the capability of the developed model in correctly classifying the new examples. Traditionally, the classification performance of binary problems can be interpreted in a confusion matrix as illustrated in Table \ref{tab1}, which can easily be extended to multi-class problems, as in our case.
One of the most commonly used measures to report the performance of classification algorithms is the average accuracy, which can be calculated as:
\begin{equation}\label{eq21}
Accuracy = \frac{TP + TN}{TP + FN + FP + TN}.
\end{equation}

However, in order to report the performance of classifiers on imbalanced datasets, other well-known metrics are used, which can be derived from Table \ref{tab1} and are formulated as:
\begin{align}\label{eq22}
Sensitivity &= \frac{TP}{TP + FN}, \\
Precision &= \frac{TP}{TP + FP},\label{eq23}\\
\label{eq24}
F-score &= (1 + \beta)\frac{Precision\times Sensitivity}{\beta^2\times Precision + Sensitivity}.
\end{align}
\noindent When $\beta = 1$, the measure in Eq.\eqref{eq24} is called the balanced F-score (F1-score), which takes both \textit{Precision} and \textit{Sensitivity} into account equally. 

\begin{table}[hbt!]
\centering
\caption{Confusion matrix.}
\label{table1}
\setlength{\tabcolsep}{3pt}
\begin{tabular}{lcc}
\hline\hline
 & \textbf{Predicted positive} & \textbf{Predicted negative}  \\
\hline
\textbf{Actual positive} & True positive (TP) &  False negative (FN)\\
\textbf{Actual negative} & False positive (FP) &  True negative (TN)\\
\hline\hline
\end{tabular}
\label{tab1}
\end{table}

Another commonly used qualitative and quantitative metric is the receiver operating characteristics (ROC), which is defined as the ratio between TP rate (Eq. \eqref{eq23}) and FP rate ($FP/(FP + TN)$) \cite{r40}. This method graphically visualises the trade-off between TP rate and FP rate. In case of multi-class classification, each curve actually evaluate the target class versus all the other classes. Beside the curves corresponding to each class, macro- and micro-average curves can also be plotted. The macro- and micro-average of precision and sensitivity are computed as follows \cite{a7}: 

\begin{align}\label{eq25}
Precision_{micro}&= \frac{\sum_{N_c} TP_{N_c}}{\sum_{N_c} TP_{N_c} + \sum_{N_c} FP_{N_c}}, \\
Sensitivity_{micro} &= \frac{\sum_{N_c} TP_{N_c}}{\sum_{N_c} TP_{N_c} + \sum_{N_c} FN_{N_c}},\\
Precision_{macro} &= \frac{\sum_{N_c} Precision_{N_c}}{{N_c}},\\
Sensitivity_{macro} &= \frac{\sum_{N_c} Sensitivity_{N_c}}{{N_c}},
\end{align}

\noindent where $N_c$ is the number of classes as defined earlier.

Furthermore, the area under the curve (AUC) of the ROC curves can be calculated as a scalar metric to evaluate the classification performance. The higher the AUC value, the better the classifier is. 

In the following section, the performance of the proposed DENS-ECG  model is evaluated using the defined classification metrics. It should be noted that F1-score, Precision (P\textsuperscript{+}), and Sensitivity (Se) are appropriate metrics for reporting the performance of classification models on imbalanced datasets. The comparison of the proposed method with two other deep learning scenarios is also reported in Section \ref{res2}. 


\subsection{QRS detection results}\label{sec3-2}
The QRS detection performance of the proposed DENS-ECG model on MITDB and QTDB databases are reported in Table \ref{tab2}. It should be noted that the results in this table are corresponding to the unseen 26 records of QTDB and the whole MITDB, which were not used during the training of the model. The proposed model achieves 99.61\%, 99.52\%, and 99.56\% in Se, P\textsuperscript{+}, and F1-score, respectively, on the well-known MITDB database for QRS detection. The model is also performed well on QTDB for QRS detection. As shown in Table \ref{tab2}, filtering the input signals improves the classification performance substantially. For example, the F1-score has been increased by more than 4\% and 7\% on MITDB and QTDB databases, respectively. In addition, the high classification performance on MITDB shows that the proposed model is well generalized, which can be used on different datasets in practice regardless of filtering technique and model parameters. The performance of DENS-ECG model is comparable with other published algorithms (Table \ref{tab5}) and the detailed discussion will be given in Section \ref{res3}.

\begin{table}[htbp]
\centering
\caption{QRS detection performance of the proposed DENS-ECG model for the MITDB and QTDB databases on the test set with and without filtering of the input signals.}
\label{table2}
\begin{tabular}{@{}lccccccc@{}}
\hline\hline
Datasets & \multicolumn{3}{c}{MITDB} && \multicolumn{3}{c}{QTDB}\\
\cmidrule{2-4}
\cmidrule{6-8}
Metrics & Se & P\textsuperscript{+} & F1-score && Se & P\textsuperscript{+} & F1-score\\
\hline
Raw signal & 96.81 & 92.01 & 95.75 && 85.54 & 99.24 & 91.89 \\
Filtered signal & 99.61 & 99.52 & 99.56 && 99.7 & 99.19 & 99.45 \\

\hline\hline
\end{tabular}
\label{tab2}
\end{table}

\subsection{ECG delineation results}
The waveforms delineation performance of DENS-ECG model on QTDB database is given in Table \ref{tab3}. The average performance of the model for detecting start, peak, and end waveforms of each four classes are reported in this table. Overall, the model performs the best on QRS detection followed by T-wave and P-wave, respectively. For example, the precision of T-wave is 5\% higher than the P-wave. However, the model sensitivity for P and T-waves detection are comparable to each other, which are more than 96.5\% with slightly more favourable for T-wave detection. The F1 score, which represents both Se and P\textsuperscript{+}, for P, QRS, T, and NW classes equals 93.01\%, 99.45\%, 96.12\%, and 98.55\%, respectively.   

As shown in Table \ref{tab3}, filtering the input signals improves DENS-ECG performance substantially. For example, the sensitivity of the model has been increased by around 20\%, 14\%, 13\%, and 14\% for P, QRS, T, and NW detection. In contrast, the improvement of model precision is not as high as sensitivity when applying filtering to the input signals.

\begin{table}[hbt!]
\centering
\caption{Average performance of the proposed DENS-ECG model on the test set for waveform delineation of four classes with and without filtering of the input signals.}
\label{table3}
\setlength{\tabcolsep}{4pt}
\begin{tabular}{@{}lcccccccccccccc@{}}
\hline\hline
Metrics & \multicolumn{4}{c}{Se} && \multicolumn{4}{c}{P\textsuperscript{+}} && \multicolumn{4}{c}{F1-score} \\
\cmidrule{2-5}
\cmidrule{7-10}
\cmidrule{12-15}
Classes & P & QRS & T & NW && P & QRS & T & NW && P & QRS & T & NW \\
\hline


Raw signal & 76.80 & 85.54 & 82.40 & 84.39 && 87.83 & 96.24 & 91.43 & 95.11 && 81.95 & 90.58 & 86.68 & 89.43 \\

Filtered signal & 96.53 & 99.7 & 96.81 & 98.75 && 89.74 & 99.19 & 95.44 & 98.36 && 93.01 & 99.45 & 96.12 & 98.55 \\

\hline\hline
\end{tabular}
\label{tab3}
\end{table}

\begin{figure}[htbp]\centering
\begin{subfigure}{.45\textwidth}
  \centering
  \includegraphics[width=0.9\linewidth]{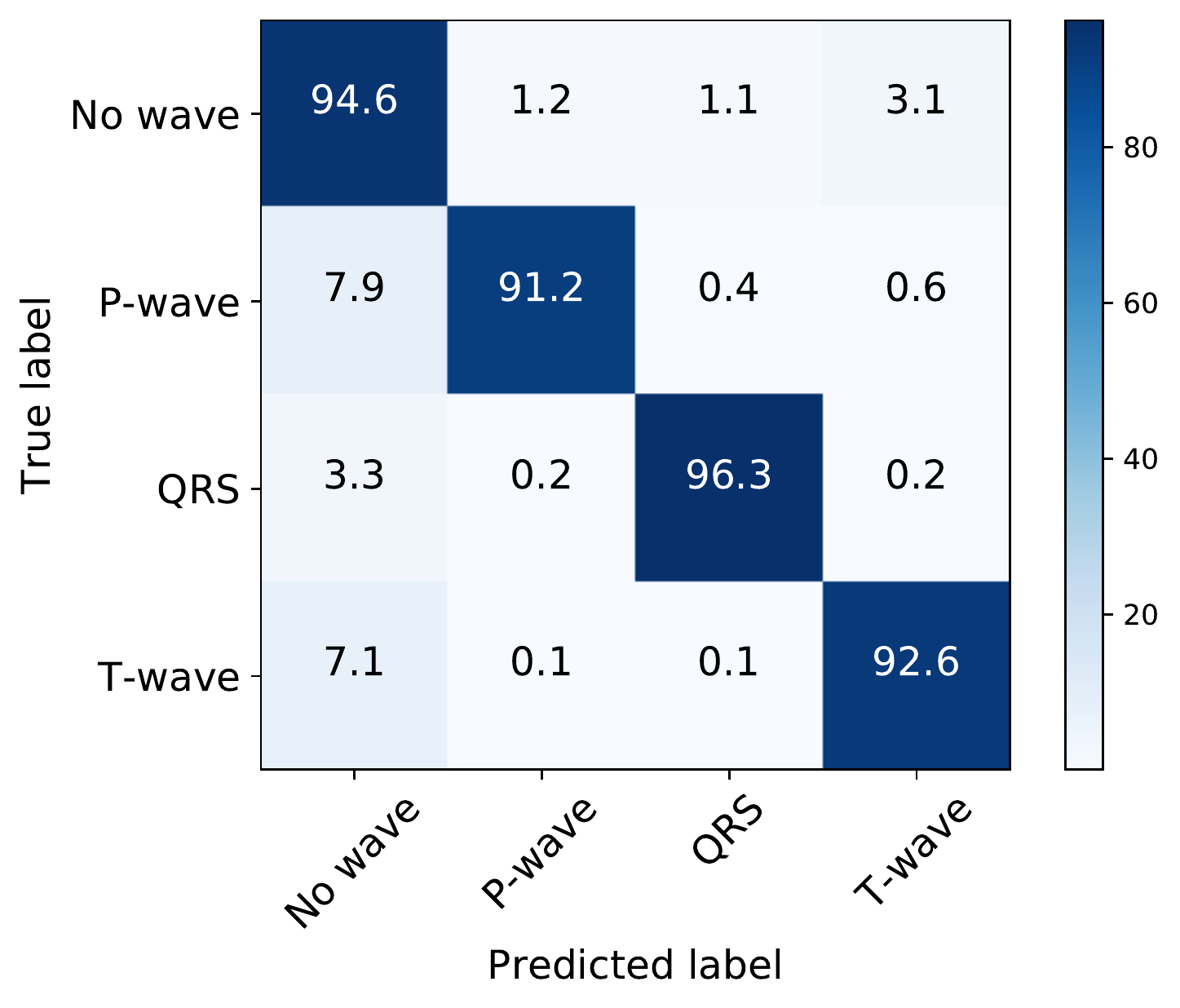}
  \caption{}
  \label{fig2:sfig1}
\end{subfigure}%
\begin{subfigure}{.45\textwidth}
  \centering
  \includegraphics[width=0.9\linewidth]{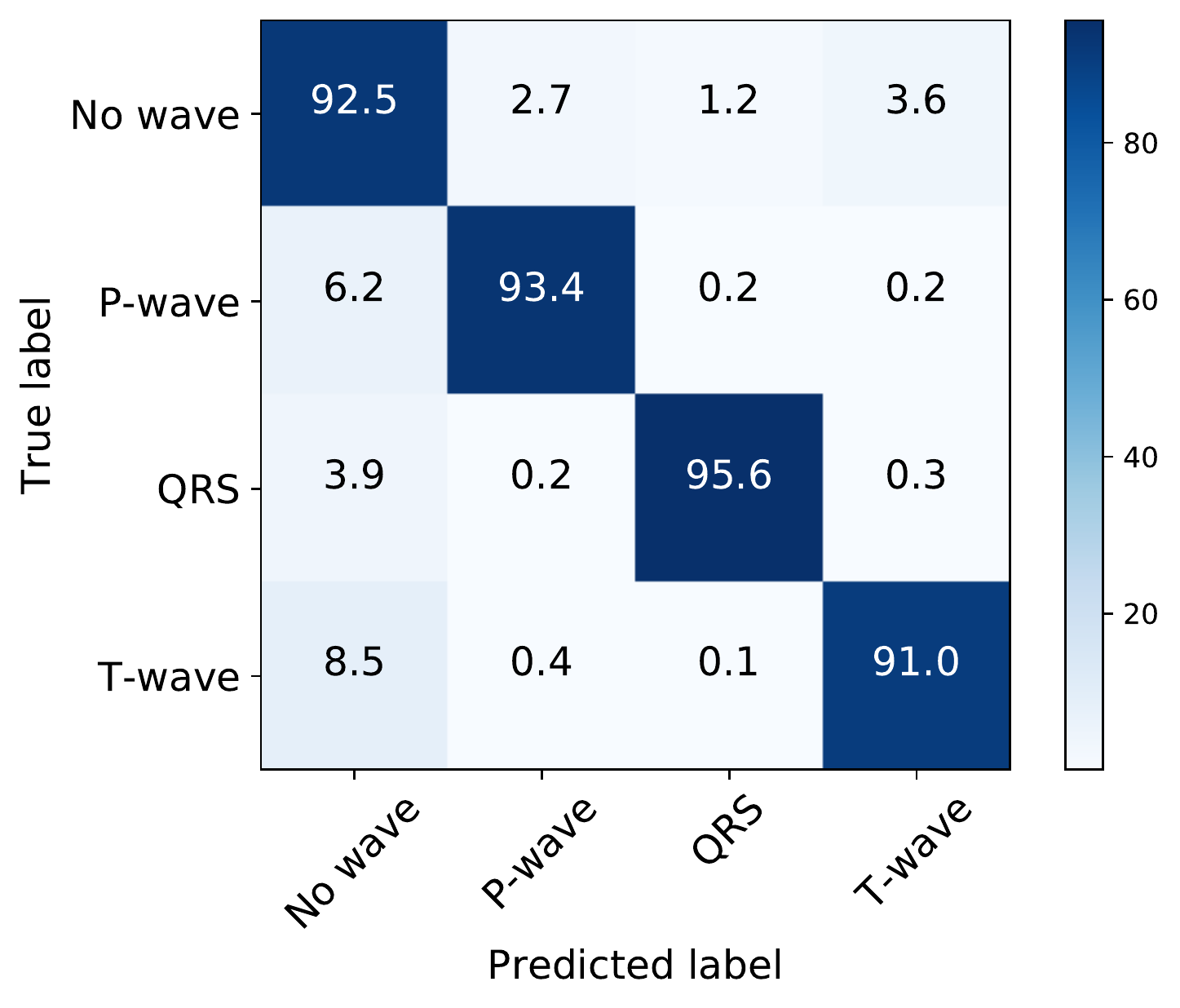}
  \caption{}
  \label{fig2:sfig2}
\end{subfigure}
\caption{Classification performance using confusion matrices for: (a) 5-fold CV and (b) test set of the proposed DENS-ECG algorithm. The numbers are in percentage.}
\label{fig2}
\end{figure}

The confusion matrices of the DENS-ECG model on the 5-fold CV and test set are shown in Figures \ref{fig2:sfig1} and \ref{fig2:sfig2}. These also confirms that the proposed DENS-ECG model performs better on QRS detection compared to other three classes. Most of the incorrect cases in all three classes (P-wave, QRS, and T-wave) are classified into NW class. In other words, the model does not make incorrect classification between the three main classes (P-wave, QRS, and T-wave). As an example, 6.2\%, 3.9\% and 8.5\% of P-wave, QRS, and T-wave classes are classified into NW class incorrectly on the test set, respectively. In addition, the small difference between 5-fold CV and test results shows that the model has been trained properly, which does not suffer from over-fitting problem.

\begin{figure}[htbp]\centering
\begin{subfigure}{.45\textwidth}
  \centering
  \includegraphics[width=0.9\linewidth]{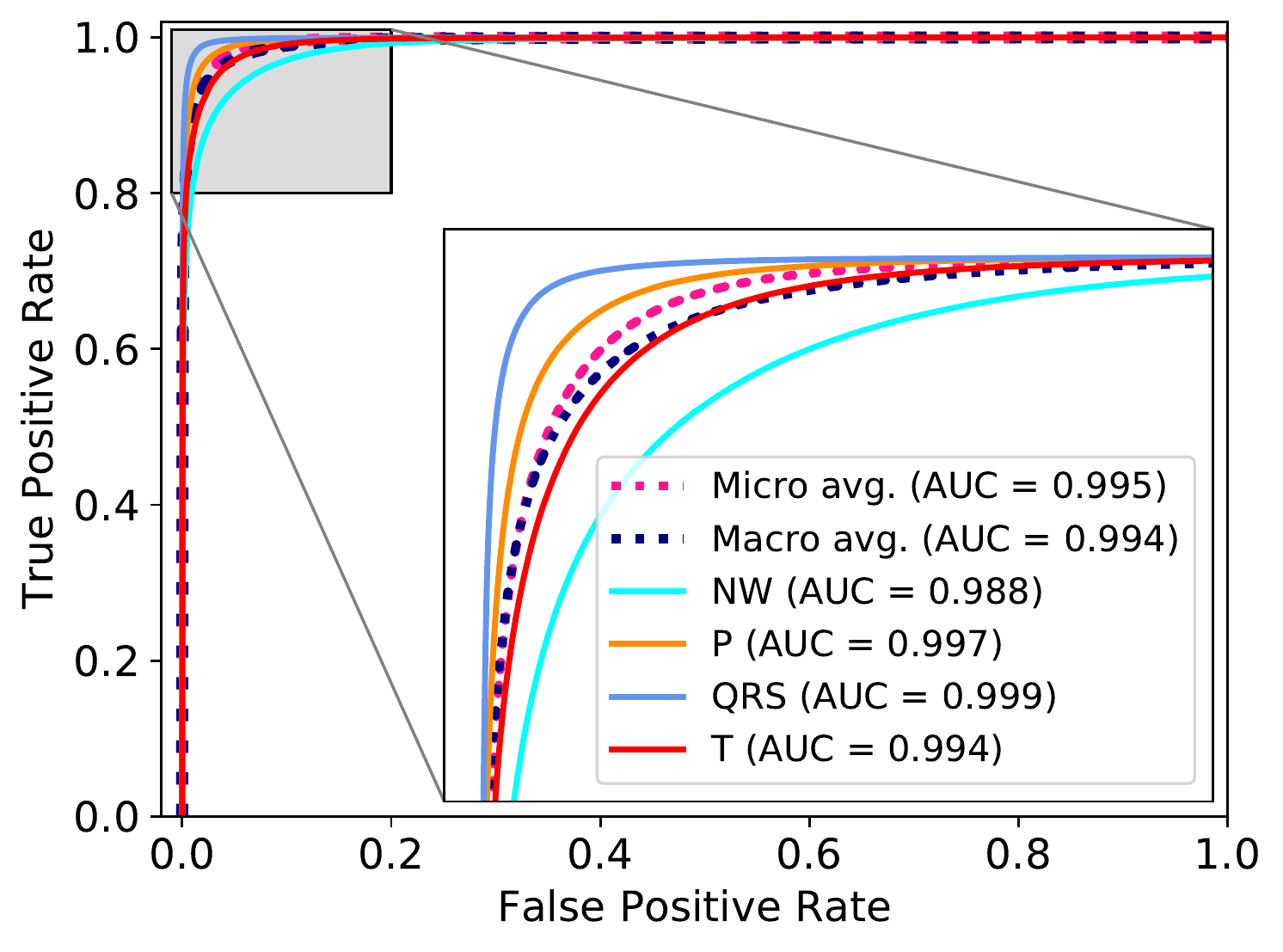}
  \caption{}
  \label{fig3:sfig1}
\end{subfigure}%
\begin{subfigure}{.45\textwidth}
  \centering
  \includegraphics[width=0.9\linewidth]{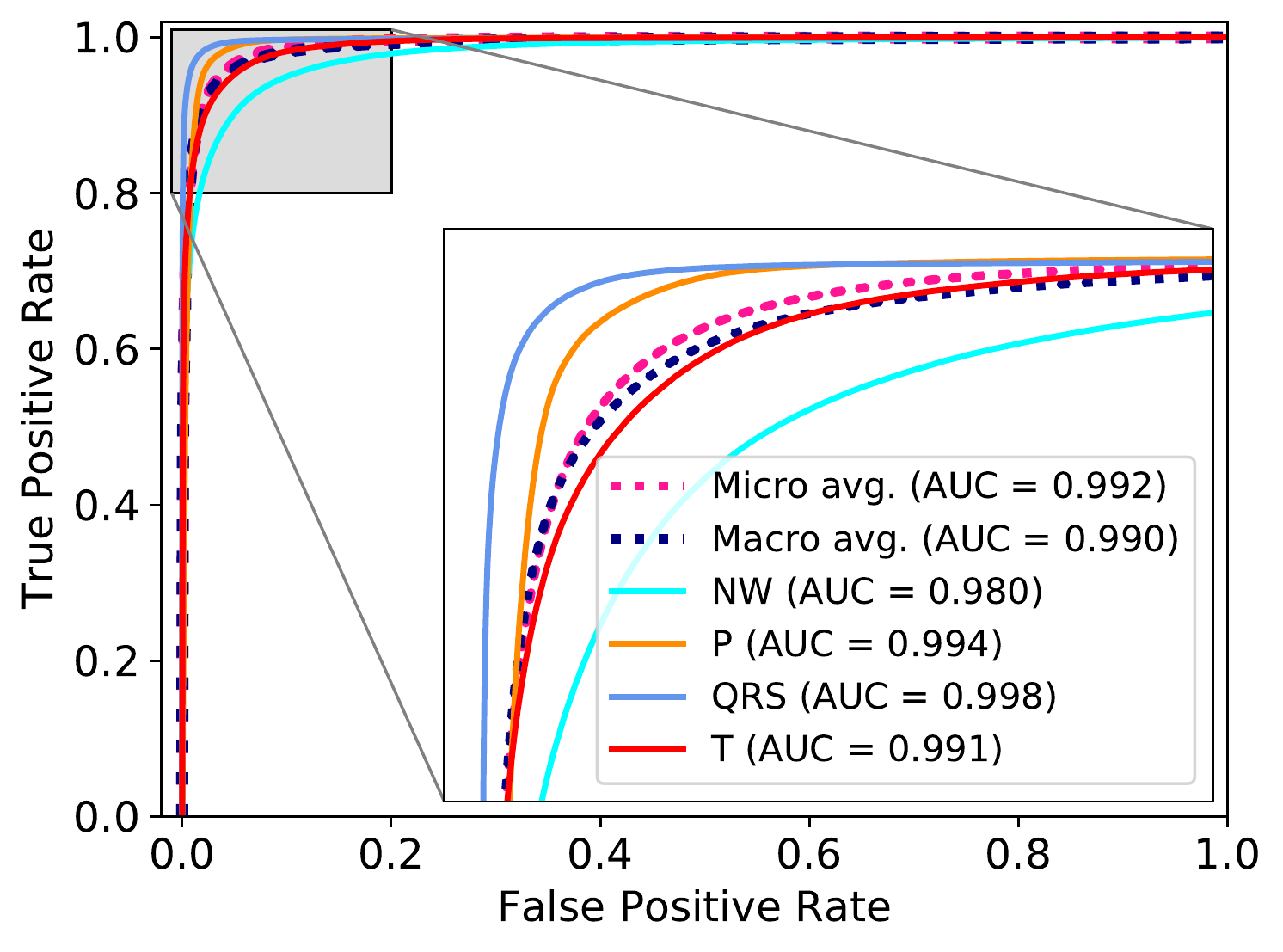}
  \caption{}
  \label{fig3:sfig2}
\end{subfigure}

\caption{ROC curves (solid lines) for four classes using the proposed deep model (DENS-ECG) together with the curves for the micro- and macro-average ROC (dashed lines). (a) 5-fold CV and (b) test set.}
\label{fig3}
\end{figure}

The ROC curve of the DENS-ECG model on the 5-fold CV and test set are plotted in Figures \ref{fig3:sfig1} and \ref{fig3:sfig2}. As shown in the zoomed areas, the model has the highest AUC on
the QRS class compared to other classes. The AUC for the micro and macro-average are 0.992 and 0.99 on the test set, respectively, showing the
promising classification performance for the DENS-ECG algorithm.

\begin{figure}[htbp]\centering
\begin{subfigure}{.48\textwidth}
  \centering
 \includegraphics[width=0.95\linewidth]{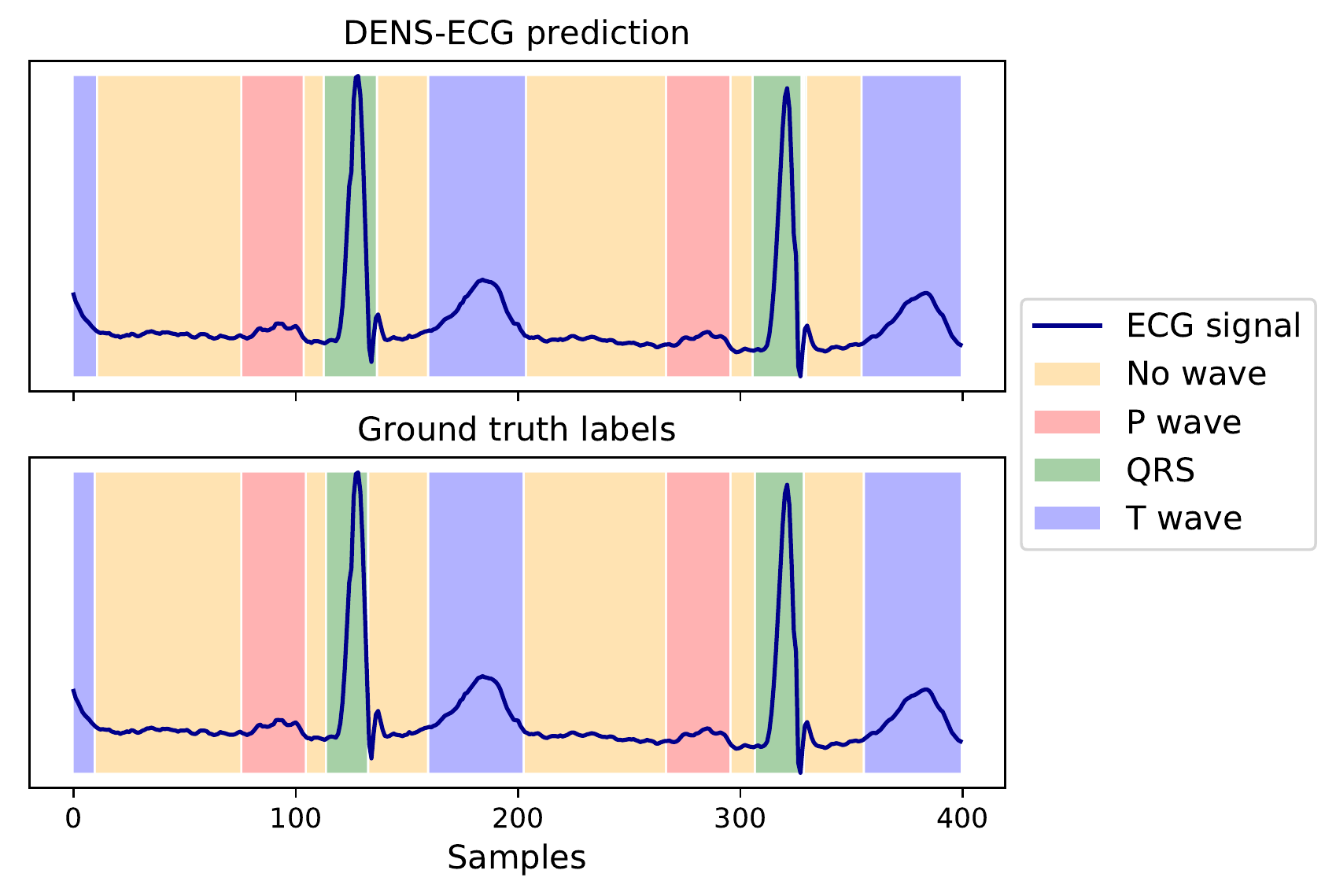}
  \caption{}
  \label{fig4:sfig1}
\end{subfigure}
\begin{subfigure}{.48\textwidth}
\centering
  \includegraphics[width=0.95\linewidth]{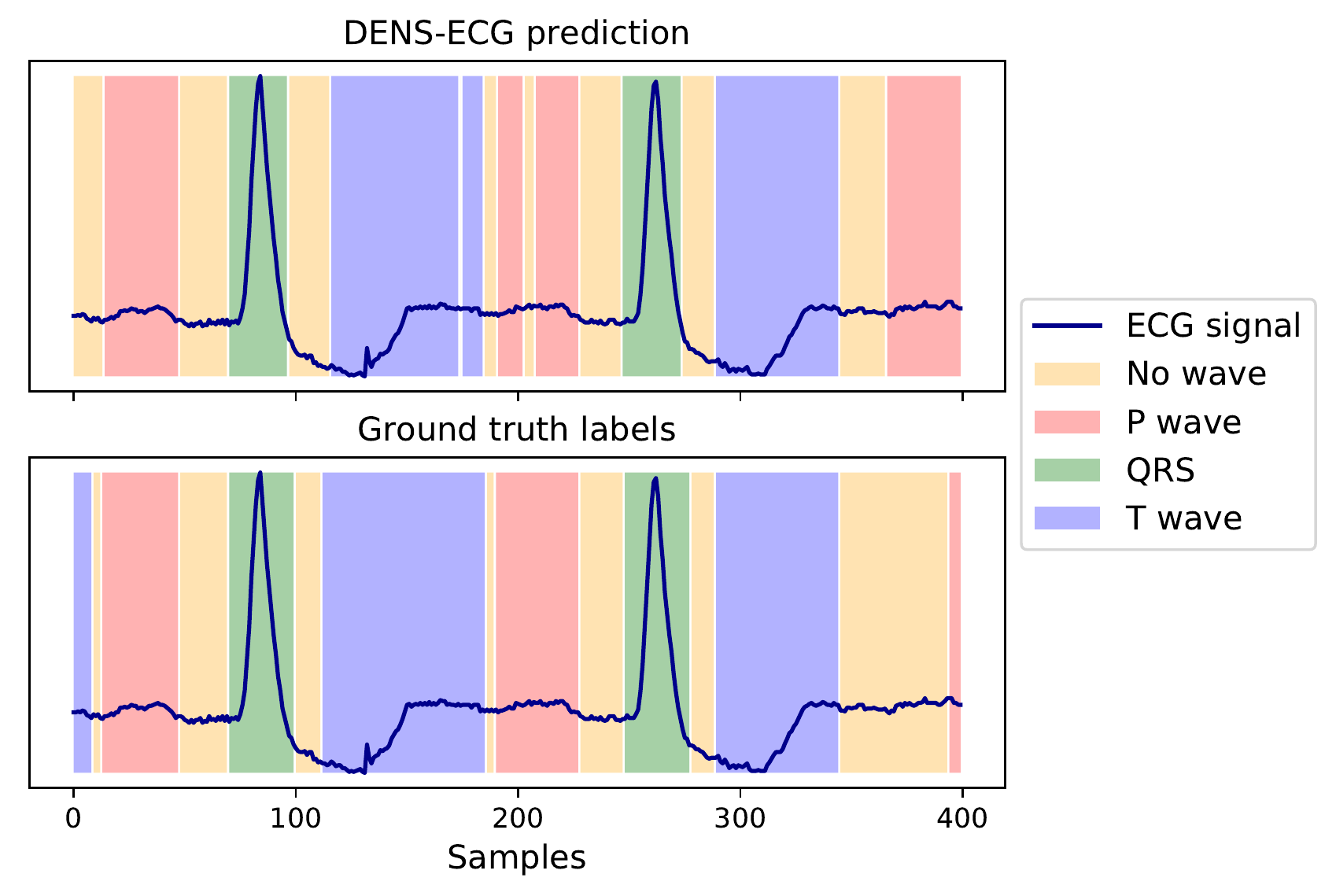}
 \caption{}
 \label{fig4:sfig2}
\end{subfigure}
\caption{Two excerpts of DENS-ECG model predictions with the corresponding detected P, QRS, T, and NW segments. The true labels (annotations) of the signals are also plotted to compare the results of the classification. (a) A prediction example with high classification rate, (b) A prediction example with some FPs and FNs.}
\label{fig4}
\end{figure}

As an example, Figures \ref{fig4:sfig1} and \ref{fig4:sfig2} show excerpts of DENS-ECG model predictions and its corresponding true annotations (labels). The developed algorithm performance for detecting P, QRS, T, and NW
segments confirms the capability of the the deep model
on delineation of these waveforms. Figure \ref{fig4:sfig1} shows an example of a nearly perfect classification in which all the four waveforms are classified correctly corresponding to the true labels. In contrast, as depicted in Figure \ref{fig4:sfig2}, in the first T-wave segment (blue strip), there are some FN predictions. For example, the prediction of the first T wave segment shows a narrow FN detection, which is followed by a correct classification. There is also a similar pattern and some FP predictions for the second P wave segment (red strip). Furthermore, there are some FN predictions for the third P wave segment. In addition, at the beginning of the prediction, there is a FP prediction for a T wave segment.

\section{Discussion}\label{sec4}
The proposed DENS-ECG model achieves a good performance on the MITDB and QTDB databases. In this section, the performance of DENS-ECG model is compared with other deep learning models as well as state-of-the-art algorithms for ECG signal delineation.

\subsection{Comparison of DENS-ECG with other deep learning approaches}\label{res2}

In this section, the performance of the proposed DENS-ECG method is compared with other deep models with different number of layers and architectures. This includes models with various number of CNN and BiLSTM layers as well as an end-to-end CNN and BiLSTM model. In all the models, the same parameters and optimisation method
as utilised in DENS-ECG are used. As given in Table \ref{tab4}, different combinations of convolutional and BiLSTM layers are examined and their corresponding performance are reported in order to compare with DENS-ECG model, which consists of three convolutional and two BiLSTM layers. It should be noted that the combination of CNN and BiLSTM layers for DENS-ECG model was inspired from our previous research work on the detection of AFIB \cite{r7}. Unlike DENS-ECG model, which uses convolutional and BiLSTM layers, the end-to-end CNN and LSTM models only use convolutional and BiLSTM layers, respectively, to extract the features and classify the waveforms. 

The end-to-end BiLSTM model consists of four layers, which are two BiLSTM layers followed by dropout and a time distributed dense layer. The number of hidden units of BiLSTM layers are the same as DENS-ECG model, which are 250 and 125, respectively. In addition, the dropout value is equal to 0.2 and there are four hidden units in the dense layer. The end-to-end CNN models consists of three convolutional layers followed by a time distributed dense layer. The number of filters in each layer are the same as DENS-ECG model, which are 32, 64, and 128, respectively.    

As reported in Table \ref{tab4}, the average F1-score (last column) for all four classes show the better performance of DENS-ECG model with three convolutional and two BiLSTM layers. Although, the performance of the model with two convolutional and two BiLSTM layers is comparable with DENS-ECG model, adding one more convolutional layer leads to the highest possible performance of the model without increasing noticeable complexity to the model. The end-to-end CNN model has the lowest performance, which achieved 71.83\% on average F1-score. On the other hand, the end-to-end BiLSTM model shows better performance than end-to-end CNN model but yet lower than other models by around 4\% in average F1-score. Table \ref{tab4} shows the effectiveness of the proposed DENS-ECG model to extract high abstract temporal features from the ECG signals in order to classify different waveforms.

\begin{table}[htbp]
\centering

\caption{Comparison of the classification performance on the test set among DENS-ECG and other deep models architectures.}
\label{table4}
\setlength{\tabcolsep}{4pt}
\begin{tabular}{@{}lcccccccccccc@{}}
\hline\hline
Metrics & \multicolumn{4}{c}{Se} && 
\multicolumn{4}{c}{F1-score} && 
\multirow{2}{*}{Avg. F1-score} \\
\cmidrule{2-5}
\cmidrule{7-10}
Classes & P & QRS & T & NW && P & QRS & T & NW && \\
\hline

1 CNN + 1 BiLSTM & 95.79 & 99.69 & 97.37 & 98.48 && 92.68 & 99.44 & 94.37 & 98.21 && 96.18\\

2 CNN + 1 BiLSTM & 94.34 & 99.68 & 99.51 & 97.83 && 92.05 & 99.45 & 95.43 & 98.12 && 96.26\\

3 CNN + 1 BiLSTM & 95.08 & 99.65 & 96.42 & 98.04 && 92.35 & 99.37 & 95.02 & 97.97 && 96.18\\

2 CNN + 2 BiLSTM & 95.48 & 99.69 & 96.58 & 98.12 && 92.64 & 99.48 & 96.44 & 98.31 && 96.72\\

3 CNN & 46.54 & 94.13 & 85.51 & 90.26 && 46.77 & 88.71 & 67.74 & 84.09 && 71.83\\

2 BiLSTM & 88.12 & 96.37 & 91.39 & 95.04 && 87.30 & 95.24 & 92.62 & 94.62 && 92.45\\

\bf{DENS-ECG} & \bf{96.53} & \bf{99.70} & \bf{96.81} & \bf{98.75}  && \bf{93.01} & \bf{99.45} & \bf{96.12} & \bf{98.55} && \bf{96.78}\\

\hline\hline
\end{tabular}
\label{tab4}
\end{table}


\begin{table}[hbt!]
\centering
\caption{Comparison of P, QRS, and T waves detection on QTDB dataset between DENS-ECG and other state-of-the-art methods. (N/R: Not Reported, N/A: Not Applicable)}
\label{table5}
\begin{tabular}{l l c c c c c c c}
\hline\hline
 Methods & Parameters & P$_{on}$ & P$_{peak}$ & P$_{end}$ & QRS$_{on}$ & QRS$_{end}$ & T$_{peak}$ & T$_{end}$ \\
\hline
\multirow{3}{*}{\cite{r43}}   & \# beats & 3194 & 3194 & 3194 & 3623 & 3623 & 3542 & 3542\\
& Se & 98.87 & 98.87 & 98.75 & 99.97 & 99.97 & 99.77 & 99.77 \\
& P\textsuperscript{+} & 91.03 & 91.03 & 91.03 & N/A & N/A & 97.79 & 97.79 \\\hline



\multirow{3}{*}{\cite{r51}}   & \# beats & 3194 & 3194 & 3194 & 3623 & 3623 & 3542 & 3542\\
& Se & N/A & N/A & N/A & N/A & N/A & 92.6 & 92.6\\
& P\textsuperscript{+} & N/A & N/A & N/A & N/A & N/A & N/R & N/R\\\hline

\multirow{3}{*}{\cite{r9}}   & \# beats & 3194 & 3194 & 3194 & 3623 & 3623 & 3542 & 3542\\
& Se & 91.2 & 91.2 & 91.2 & N/R & N/R &  93.6 &  93.6 \\
& P\textsuperscript{+} & N/R & N/R & N/R & N/A & N/A & N/R & N/R \\\hline

\multirow{3}{*}{\textbf{DENS-ECG}}    & \# beats & 761 & 761 & 761 & 851 & 851 & 834 & 834\\
& Se & 95.49 & 97.69 & 96.41 & 99.75 & 99.36 & 97.71 & 95.87 \\
& P\textsuperscript{+} & 88.77 & 90.84 & 89.06 & N/A & N/A & 96.51 &  94.43\\
\hline\hline
\end{tabular}
\label{tab5}
\end{table}

\begin{table}[hbt!]
\centering
\caption{Comparison of QRS detection on MITDB dataset among DENS-ECG and other state-of-the-art methods.}
\label{table6}
\begin{tabular}{lccccccc}
\hline\hline
 Methods & \# beats & TP & FP & FN & Err. (\%) & Se & P\textsuperscript{+} \\
\hline
\cite{r43}   & 109428 & 109208 & 153 & 220 & 0.34 & 99.80 & 99.86 \\
\cite{r54}    & 109481 & 109146 & 137 & 135 & 0.43 & 99.69 & 99.88 \\
\cite{r53}   & 109809 & 109532 & 507 & 277 & 0.71 & 99.75 & 99.54\\
\cite{r55}    & 109963 & 109522 &  545 & 441 & 0.90 & 99.60 & 99.50 \\
\textbf{DENS-ECG}    & \textbf{109494} & \textbf{109066} & \textbf{525} & \textbf{428} & \textbf{0.87} & \textbf{99.61} & \textbf{99.52} \\
\hline\hline
\end{tabular}
\label{tab6}
\end{table}

\subsection{Comparison of DENS-ECG with other state-of-the-art methods}\label{res3}
The proposed DENS-ECG model is evaluated on the test set (26 records) of QTDB database for ECG waveform delineation. The model is also compared with other ECG delineation algorithms as given in Table \ref{tab5}. 

It should be taken into account that the number of heartbeats (annotations) used for evaluating DENS-ECG in Table \ref{tab5} is less than other methods since only 26 out of 105 records are used as test set. Although, the performance of DENS-ECG model is not as high as the algorithm reported in \cite{r43}, it outperforms other algorithms published in \cite{r51} and \cite{r9} in delineation of P- and T-wave. As it can be seen from Table \ref{tab5}, the capability of DENS-ECG model in QRS\textsubscript{on} and QRS\textsubscript{end} detection is comparable with other models with sensitivity equals to 99.75\% and 99.36\% for QRS\textsubscript{on} and QRS\textsubscript{end}, respectively. While the first three models in Table \ref{tab5} outperform DENS-ECG especially in P\textsubscript{on}, P\textsubscript{end}, and T\textsubscript{end} detection. 

However, the proposed DENS-ECG model can be considered as a fully automated data-driven method, which does need minimum parameters tuning to be used in practice. Furthermore, the model is generalized enough to be applied on different datasets in practice. As reported in Section \ref{sec3-2} and Table \ref{tab6}, the model performance on MITDB dataset shows its capability in analyzing unseen datasets. In terms of required preprocessing step in DENS-ECG model, it should be also noted that the filtering of MITDB dataset is identical to what was used for training the model on QTDB dataset. 

As shown in Table \ref{tab6}, the results of DENS-ECG model in QRS detection is comparable with other algorithms (Se=99.61\% and P\textsuperscript{+}=99.52\%). The wavelet-based model introduced by \cite{r43} has the highest performance for both sensitivity and precision (Se=99.8\% and P\textsuperscript{+}=99.86\%) compared to other algorithms in Table \ref{tab6} followed by the model proposed by \cite{r54}. From Table \ref{tab6}, the performance of proposed DENS-ECG model is comparable with the well-known \cite{r53} algorithm for QRS detection. Furthermore, the model performs slightly better than QRS detection algorithms proposed in \cite{r55}.

\section{Conclusion}\label{sec5}
One of the most challenging tasks in ECG waveform delineation has been the detection of P, QRS, and T waves.  In this paper, a deep learning approach, named as the DENS-ECG, that combines the CNN-LSTM networks was proposed to predict the ECG waveforms. The laborious feature extraction step was omitted and the filtered ECG segments were directly used as inputs for training the model. The deep CNN-LSTM network was utilised to extract highly abstract temporal features from 1D ECG signals. These features were then used to classify four different classes (P, QRS, T and NW). The model was trained using stratified 5-fold cross validation technique. Finally, the trained model was tested on a completely unseen test sets to evaluate the performance of the classification algorithms. The outputs of the model at each time stamp were the posterior probabilities assigned to the four classes. The proposed model shows a high performance on the test sets with an average F1-score of 99.56 and 96.78 on the MITDB and QTDB datasets, respectively. The efficacy of the proposed DENS-ECG model in detecting ECG waveforms provides us with the opportunity to use this algorithm \textit{in house} by cardiologists to analyze ECG recordings in order to diagnose cardiac arrhythmias such as AFIB.

\section*{Acknowledgements}

The authors would like to thank the support given by the
Innovation Fund Denmark (IFD) under Project No. 6153-00009B.

\bibliographystyle{unsrt}  
\bibliography{main.bib}  






\end{document}